\documentclass[letterpaper, 10 pt, conference]{ieeeconf}
\usepackage{times}
\usepackage[pdftex]{graphicx}
\usepackage{algorithm}
\usepackage[bottom]{footmisc}
\usepackage{algpseudocode}
\usepackage{amsmath,amssymb,amsopn,amstext,amsfonts}
\usepackage{cancel}
\usepackage[space]{cite}
\usepackage{pdfsync}
\usepackage{balance}
\usepackage{color}
\usepackage{mathtools}
\usepackage{bm}

\usepackage{diagbox}
\usepackage{float}
\usepackage{epstopdf}
\usepackage{pifont}
\usepackage{fixltx2e}
\usepackage{amsmath}
\usepackage{multirow}
\usepackage{booktabs}
\usepackage{url}
\usepackage{svg}
\usepackage{threeparttable}
\usepackage[linkcolor=black,citecolor=black,urlcolor=black,colorlinks=true]{hyperref}
\usepackage{subcaption}
\algnewcommand{\algorithmicgoto}{\textbf{go to line}}
\algnewcommand{\Goto}[1]{\algorithmicgoto~\ref{#1}}

\DeclareGraphicsExtensions{.png,.jpg,.eps,.pdf}
\IEEEoverridecommandlockouts
\makeatletter
\def\endthebibliography{%
 \def\@noitemerr{\@latex@warning{Empty `thebibliography' environment}}%
 \endlist
}

\title{\LARGE \bf
Continuous Implicit SDF Based \\ Any-shape Robot Trajectory Optimization 
}

\author{Tingrui Zhang $^{\dag \,}$, 
   Jingping Wang $^{\dag \,}$, 
 Chao Xu, Alan Gao, and Fei Gao
 \thanks{\textbf{${\dag}$ Equal contribution.}}        
 \thanks{All authors are from Zhejiang University, Hangzhou, 310013, China and the Huzhou Institute of Zhejiang University, Huzhou, 313000, China.} 
 \thanks{Email:{\tt\footnotesize \{tingruizhang, 22232111, fgaoaa\}@zju.edu.cn}}
 \thanks{Corresponding Author: Fei Gao}
}

\begin{document}
\makeatletter
 \let\@oldmaketitle\@maketitle
 \renewcommand{\@maketitle}{\@oldmaketitle
  \includegraphics[width=0.91\linewidth]
  {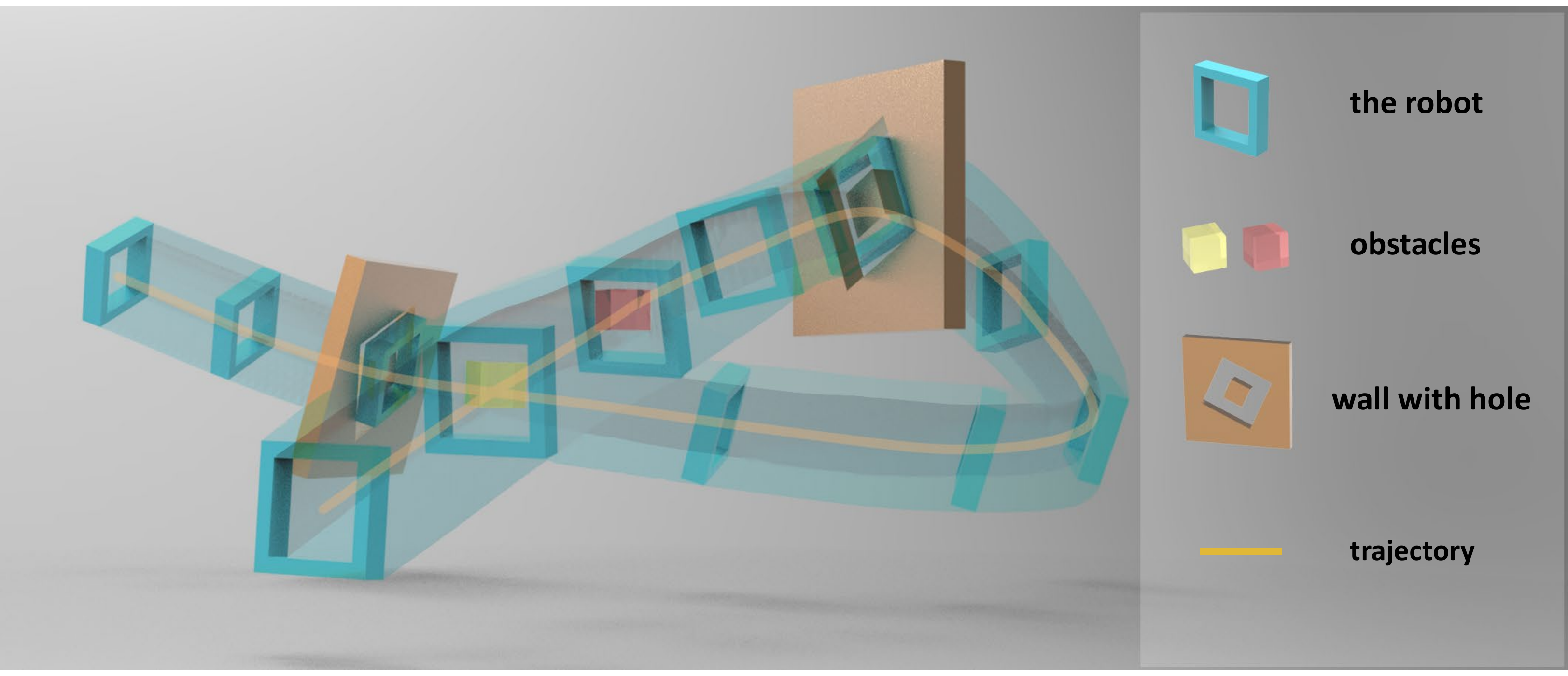}
  \centering
  \captionsetup{font={small}}
  \captionof{figure}{ \label{fig: top}
By integrating our proposed implicit SDF into the trajectory optimization, the swept volume of the box robot can avoid obstacles. Taking advantage of our formulation, our algorithm applies to robots of any shape and achieves continuous collision avoidance.
  }
  \vspace{0.0cm} 
 }
 \makeatother
 \maketitle
\setcounter{figure}{1}
    \thispagestyle{empty}
    \pagestyle{empty}

\begin{abstract}
Optimization-based trajectory generation methods are widely used in whole-body planning for robots.
However, existing work either oversimplifies the robot's geometry and environment representation, resulting in a conservative trajectory, or suffers from a huge overhead in maintaining additional information such as the Signed Distance Field (SDF). To bridge the gap, 
we consider the robot as an implicit function, with its surface boundary represented by the zero-level set of its SDF. Based on this, we further employ another implicit function to lazily compute the signed distance to the swept volume generated by the robot and its trajectory.
The computation is efficient by exploiting continuity in space-time, and the implicit function guarantees precise and continuous collision evaluation even for nonconvex robots with complex surfaces. Furthermore, we propose a trajectory optimization pipeline applicable to the implicit SDF.
Simulation and real-world experiments validate the high performance of our approach for arbitrarily shaped robot trajectory optimization.
\end{abstract}

\section{Introduction}
\label{sec:Introduction}
Whole-body planning is critical for robots in dense environments. To this problem,
optimization-based trajectory generation approaches are effective and have received much attention, with critical considerations including the shape of robots and leveraging environment information from maps.

 Two main methods exist for representing a robot's shape: enclosing it in  simple geometric shapes, such as ellipsoids, cylinders, polyhedrons\cite{zhang2022generalized,han2021fast} or using surface samples to represent its geometry\cite{5653520}. The former lacks precision and the represented shape is often larger than the actual size of the robot, resulting in conservative trajectories. The latter is limited by the resolution, potentially resulting in collision risks at low resolution or complex representation at high resolution. For map representations, optimization-based trajectory planning methods typically require additional information to be recorded in maps to construct safety constraints, such as  SDFs\cite{oleynikova2016signed} and safe corridors\cite{guo2022dynamic} of the environment. However, this introduces extra computational and memory overhead. Furthermore, SDFs with a low resolution cannot represent the complex environment precisely, thus adversely affecting the robot's trajectory planning in dense environments. Safe corridors sacrifice a lot of solution space.

 In conclusion, existing methods suffer from the  following two problems:
\begin{itemize}
    \item [1)] It is hard to model a robot in a general and efficient way for whole-body planning since the robot's shape may be complex e.g., non-convex or even changing over time.
    \item[2)] Trajectory optimization usually requires additional information such as SDFs or safe corridors to construct safety constraints, with their corresponding drawbacks. 
\end{itemize}
Based on the above issues, we find that there is still no unified framework that can effectively handle the trajectory generation for arbitrarily shaped robots.

To bridge this gap, we propose a novel approach to whole-body trajectory generation using a continuous implicit SDF representation. Our approach does not rely on simplification using simple geometric shapes or surface sampling to represent the robot. Instead, we use the original geometry to achieve accurate modeling. The key insight is that the surface boundary of any robot is represented by the zero-level set of its SDF. Moreover, our approach is applicable to different environment representations. We don't need to compute or store SDF for the whole map, is independent of resolution, and also enjoys much more solution space due to our formulation. We combine an implicit SDF representation with the concept of swept volume in computational geometry inspired by \cite{blackmore1992analysis,sellan2021swept,614295}.  Based on this, we formulate a continuous implicit function and implement an optimization-based pipeline for any-shape robot trajectory generation.

Swept volume refers to the three-dimensional space occupied by an object as it moves through its entire range of motion. 
Our approach can evaluate the safety of the swept volume of an arbitrarily shaped robot and ensure that the interior of the swept volume remains free from contact with any obstacles, thus achieving the generation of collision-free trajectories. The SDF is  implicitly constructed based on the swept volume and lazily evaluated only at some interested obstacle points with known coordinates.
Therefore, it is applicable to different environment representations, such as point cloud maps, feature maps, and grid maps, as long as the coordinates of the obstacles are known. Moreover, no additional information from maps is required. To verify the feasibility and capability of the proposed approach, we perform simulation and real-world experiments on a quadrotor platform. The proposed algorithm is less conservative and offers a wider solution space in optimization, and does not require any complex environment representation. Moreover, benefiting from the
continuous implicit SDF, our method can achieve continuous collision  avoidance.

We summarize our contributions as follows:
\begin{itemize}
\item [1)]We consider a robot as an implicit function and propose an algorithm to efficiently obtain the SDF of a swept volume  by exploiting the continuity in space-time.
\item [2)]We propose an optimization-based planning pipeline for any-shape robots, which is based on the continuous implicit SDF and enables continuous collision avoidance.
\item [3)]We will open source our algorithm\footnote{\url{https://github.com/ZJU-FAST-Lab/Implicit-SDF-Planner}} for the reference of the community.
\end{itemize}

\begin{figure}[!th] 
\centering
\vspace{0.5cm}
{\includegraphics[width=1.0\columnwidth]{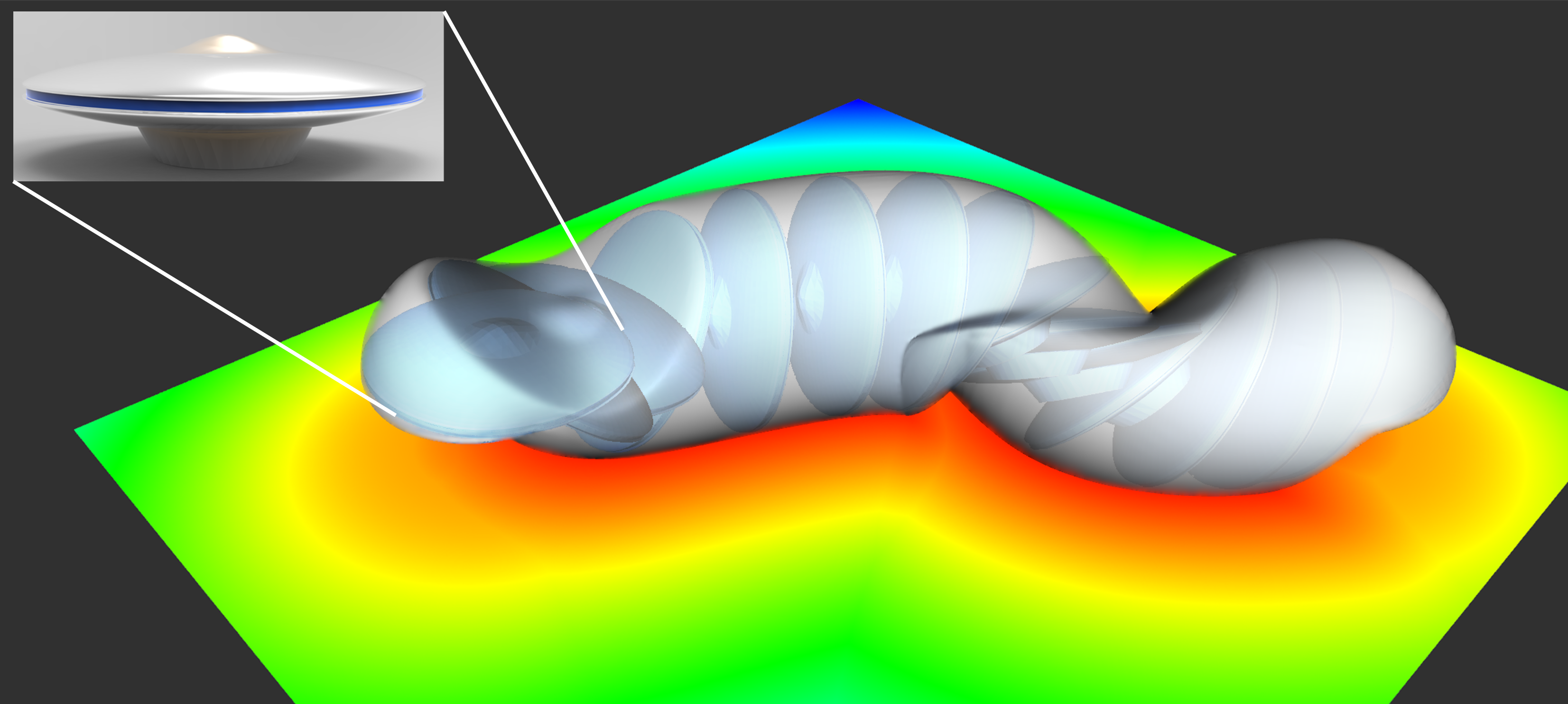}}
\captionsetup{font={small}}
\caption{ \label{fig: SDFquery} The swept volume generated by a UFO robot moving along a trajectory with its corresponding SDF is shown here. For ease of visualization, only the horizontal plane at z=0 is rendered and the signed distance is displaced using rainbow colors.}
\vspace{-0.8cm}
\end{figure} 

\section{Related Works}
\label{sec:RelatedWork}
\subsection{Geometric Shape Representations For Motion Planning}
\label{subsec:Whole-Body}
Geometric representations and computations play an important role in robotics \cite{toth2017handbook}, especially for whole-body planning. Most research focuses on using some convex geometric shapes such as ellipsoids, polyhedrons, or cylinders to model  configuration space or robots for efficient performance, but this sacrifices some solution space.

To represent configuration space, using sets of polyhedrons to construct safe corridors has been widely adopted by some work \cite{9543598,ding2019safe,wang2022geometrically}. This representation can be utilized to impose safety constraints for robots in motion planning. 
However, this simplification may be too conservative for some robots with non-convex shapes due to the introduced gap in collision evaluation.   
Recently, Tracy et al.\cite{tracy2022differentiable} use ellipsoids, capsules, boxes, and
their combinations as collision primitives to approximate complicated rigid bodies. By formulating a distance minimization problem and using its corresponding derivatives, whole-body planning is achievable. However, this method introduces approximation errors and the gradient computation is cumbersome and relatively expensive. 
Similarly, Wang et al. \cite{Wang2022ALA} model obstacles and robots as polyhedrons and use scale optimization to achieve collision evaluation and whole-body planning. While being exact and having no gap, this method is not suitable for  non-convex robots or obstacles. 

\begin{figure}[!thb]  
\centering
\vspace{0cm}
{\includegraphics[width=0.95\columnwidth]{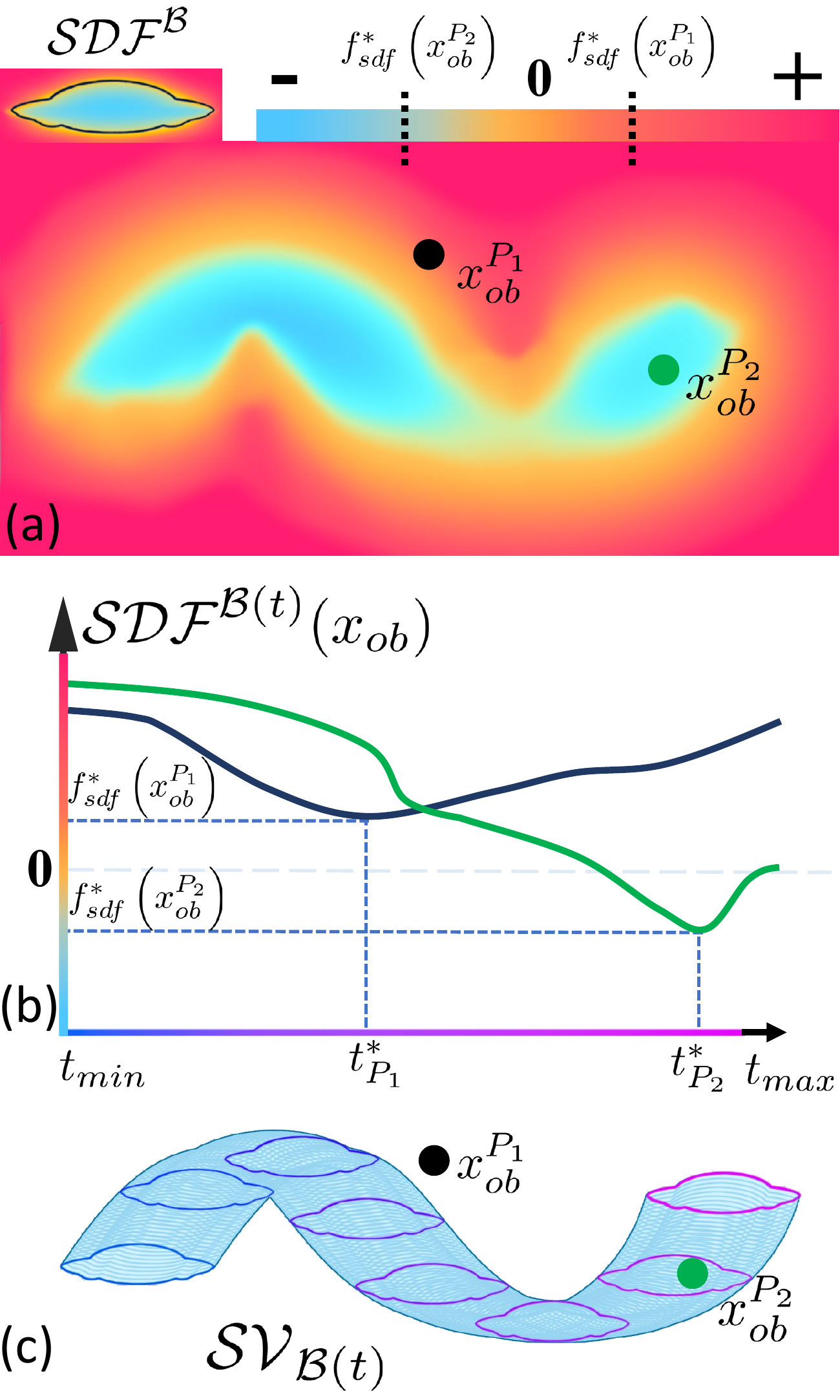}}
\captionsetup{font={small}}
\caption{ \label{fig: pic2}$\mathcal{SV}_{\mathcal{B}(t)}$ of a UFO robot is shown in Fig.c with its corresponding SDF shown in  Fig.a. Fig.b shows the value of $\mathcal{SDF}^{\mathcal{B}(t)}(\bm{x}_{ob})$ in the time domain when a query point $\bm{x}_{ob}$ is given. Points $P_1$ and $P_2$ along with their respective $f_{sdf}^*$ and $t^*$ values are shown here.}
\vspace{-0.3cm}
\end{figure} 

\subsection{SDF Representations in Robotics}
\label{subsec: SDF}
The SDF provides useful information about the distance between a robot and nearby obstacles, which has great potential for motion planning.
Most work generates the SDF on pre-constructed maps. For example, the SDF can be constructed on grid maps using the method in \cite{felzenszwalb2012distance}. \cite{oleynikova2017voxblox} and \cite{han2019fiesta} proposed a kind of fast incremental SDF construction method that can be applied in a dynamic environment.
However, in robotics, such approaches have several drawbacks. First, constructing the SDF for an entire map consumes a significant amount of memory resources, making it a challenging task in terms of memory overhead. Second, such approaches require a specific resolution to strike a balance between system overhead and accuracy, which limits their applications in complex and complicated environments. Third, the SDF of a entire map is considered redundant, which is discussed in detail in \cite{zhou2020ego}. Therefore, in robotics, lazy querying is much more attractive to minimize unnecessary computational or memory overhead.

\section{Continuous Implicit SDF Generation}
A trajectory in $\mathbb{SE}(3)$ consists of a position trajectory $p(t)$ and an attitude trajectory $R(t)$, where $p \in \mathbb{R}^3$ and $R \in \mathbb{SO}(3)$.
Given a robot representation $\mathcal{B}$, its configuration in $\mathbb{SE}(3)$ can be computed as  $\mathcal{B}(t) = R(t)\mathcal{B} + p(t)$. The swept volume generated by the motion of $\mathcal{B}$  during its rigid transformation can be denoted as $\mathcal{SV}_{\mathcal{B}(t)}$.

As mentioned in chapter \ref{sec:Introduction}, the idea of our algorithm is to guarantee that the $\mathcal{SV}_{\mathcal{B}(t)}$ does not collide with obstacles. Therefore, a metric is needed to evaluate the safety of the $\mathcal{SV}_{\mathcal{B}(t)}$.
Signed distance is a commonly used safety metric in robotics and is very easy to be applied in trajectory optimization.
In this chapter, we will effectively compute the signed distance of $\mathcal{SV}_{\mathcal{B}(t)}$ by exploiting the continuity in space-time. The lazily computed signed distance will be used for trajectory optimization in chapter \ref{sec: Trajectory Generation} to achieve safe trajectory generation for any-shape robots.

\subsection{Implicit SDF Representation of Robots}
\label{implicitsdf}
Recall that the surface boundary of any robot is represented by the zero-level set of its SDF. Thus we use it as an implicit continuous function $\mathcal{SDF}^\mathcal{B}:\mathbb{R}^{3}\rightarrow \mathbb{R}$ to represent an arbitrarily shaped robot that takes a negative value inside $\mathcal{B}$ $\left(\mathcal{SDF}^\mathcal{B} (\bm{x}_{ob})<0: \bm{x}_{ob} \in \mathcal{B}\right)$.
The implementation of $\mathcal{SDF}^\mathcal{B}$ simply relies on off-the-shelf libraries.

In computational geometry, the method of using triangular meshes is the most general and mature way to represent an arbitrary shape\cite{botsch2006geometric}.
For robot geometry representation, triangular meshes are used here to achieve accurate shape modeling.
The implicit SDF is realized by using the winding number signed distance field\cite{xu2020signed}. Off-the-shelf algorithms such as generalized winding number\cite{barill2018fast} and semi-general purpose axis-aligned bounding box hierarchy within the LIBIGL\footnote{\url{https://libigl.github.io/}} allow the computation of an implicit SDF along with its gradient. With this representation, given a robot $\mathcal{B}$ of any shape, the signed distance $\mathcal{SDF}^\mathcal{B}(x)$ and the gradient $\nabla{\mathcal{SDF}^{\mathcal{B}}}\big|_{x}$ at any query point $x$  can be computed efficiently with little overhead. 

\subsection{Implicit SDF Representation of Swept Volume }
\label{svsdf}
The time-invariant function $\mathcal{SDF}^\mathcal{B}$ transforms into a time-varying function as a result of the robot's motion:
 \begin{flalign}
  \label{SDFquery} 
&f_{sdf}\!\left( \bm{x}_{ob},t\right)\!=\!\mathcal{SDF}^{\mathcal{B}(t)}(\bm{x}_{ob})\!=\!\mathcal{SDF}^{R(t) \mathcal{B}+p(t)}(\bm{x}_{ob}).    
 \end{flalign}
Based on the relativity of motion, equation (\ref{SDFquery}) can be rewritten as:
 \begin{equation}
 \label{SDFquery2}
     f_{sdf}\left( \bm{x}_{ob},t\right)=\mathcal{SDF^{\mathcal{B}}}(R^{-1}(t)(\bm{x}_{ob}-p(t)),
 \end{equation}
 and its derivative with respect to $t$ is:
\begin{flalign}
\label{argdot} 
  &{\dot f_{s d f}}\big|_{\bm{x}_{ob}} \!=\!({\nabla{\mathcal{SDF}^{\mathcal{B}}}\big|_{\bm{x}_{rel}}})^{T}(R^{-1} \dot R R^{-1}(p-\bm{x}_{ob})\!-\!R^{-1} v).
\end{flalign}
The term $\bm{x}_{rel}$ refers to $R^{-1}(\bm{x}_{ob}-p)$ and the symbol $v$ represents the velocity, namely $\dot{p}$.
As Figure.\ref{fig: pic2} shows, for any given query point $\bm{x}_{ob}$, $\mathcal{SDF}^{ \mathcal{B}(t)}(\bm{x}_{ob})$ is a time-variant function due to the motion of $\mathcal{B}$. 
Intuitively, if $f_{sdf}(\bm{x}_{ob}, t)$ reaches its minimum value in the time domain and the corresponding moment is $t^*$, then $\mathcal{SDF}^{ \mathcal{B}(t^*)}(\bm{x}_{ob})$ is the signed distance of $\bm{x}_{ob}$ to the $\mathcal{SV}_{\mathcal{B}(t)}$. Assuming that $p(t)$ and $R(t)$ are continuous then $\mathcal{SDF}^{ \mathcal{B}(t)}(\bm{x}_{ob})$ enjoys continuity in space-time, which makes it easy to obtain the minimum value of $f_{sdf}$ by some numerical methods. 

Denote the signed distance of $\bm{x}_{ob}$ with respect to $\mathcal{SV}_{\mathcal{B}(t)}$, namely the minimum value of $f_{sdf}$ as follows:
\begin{equation}
 \label{implict}
     f_{s d f}^{*} \left( \bm{x}_{ob}\right) \triangleq \min _{t \in\left[t_{min}, t_{max}\right]} \mathcal{SDF^{\mathcal{B}}}(R^{-1}(t)(\bm{x}_{ob}-p(t)),
 \end{equation}
where the associated argmin time is denoted as follows:
 \begin{flalign}
\setlength\abovedisplayskip{5pt}
\setlength\belowdisplayskip{5pt}
 \label{arg}
     &t^{*}(\bm{x}_{ob}) \! \triangleq \! \underset{t \in\left[t_{\min ,} t_{\max }\right]}{\operatorname{argmin}} \! \mathcal{SDF}^{\mathcal{B}} \! \left(R^{-1}(t)(\bm{x}_{ob}\!-\!p(t))\right)\!.
 \end{flalign}
 
\begin{figure}[!t]          
\centering
{\includegraphics[width=0.95\columnwidth]{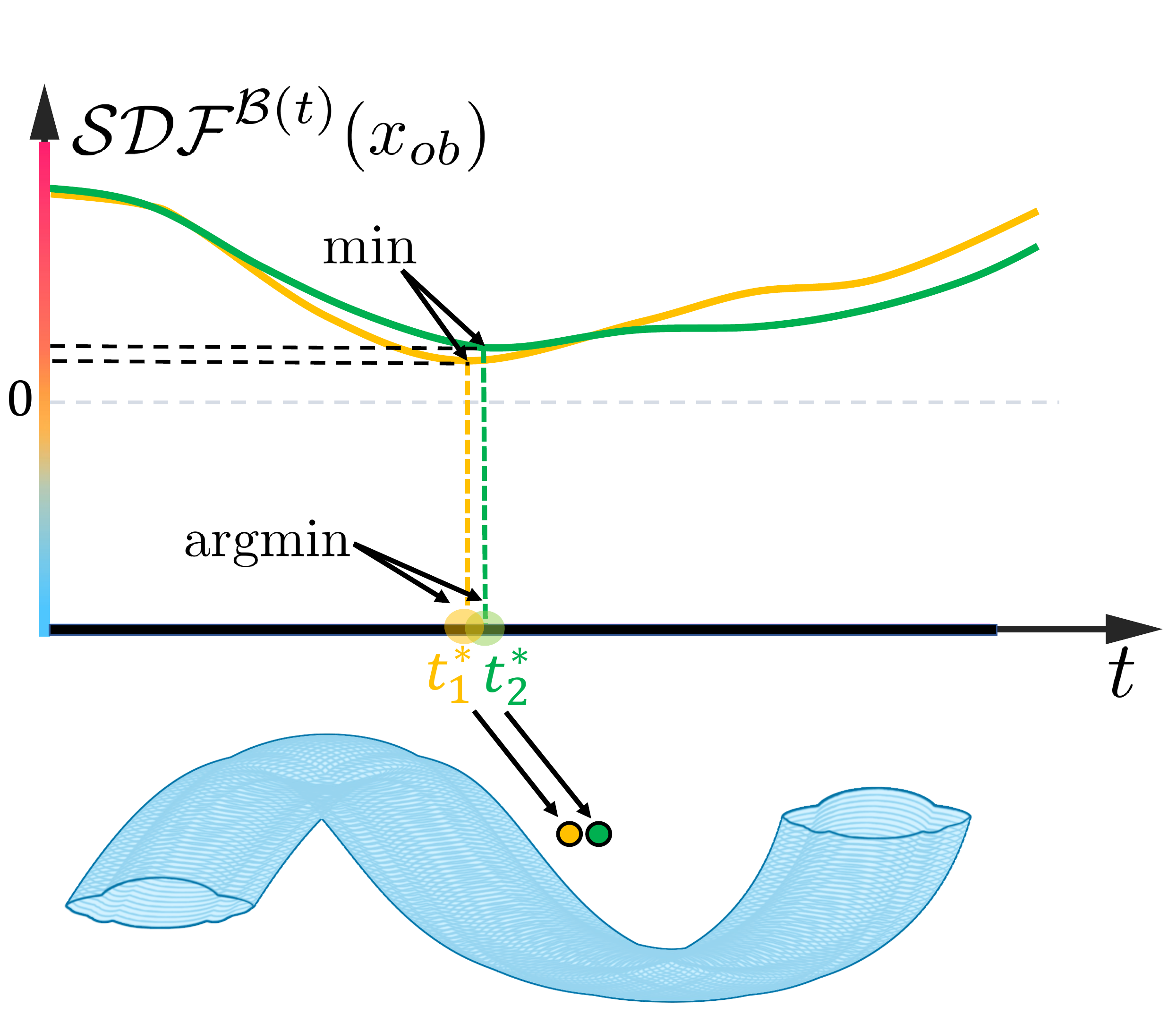}}
\captionsetup{font={small}}
\caption{ \label{fig: continuation}  The function $\mathcal{SDF}^{\mathcal{B}}\left(R^{-1}(t)(\bm{x}_{ob}-p(t))\right)$ exhibits continuity in both the space and time domains. Furthermore, the argmin $t^*$ exhibits piecewise continuity. The proximity of the yellow and green points in space is also reflected in the proximity of their respective argmin values, $t^{*}_1$ and $t^{*}_2$ in time as shown here.}
\vspace{-1.0cm}
\end{figure}

We employ a continuation technique to tackle the problem in equation (\ref{implict}). Rather than computing the minimum value directly, we focus on its associated argmin since we have observed that $t^*$ exhibits piecewise continuity over the spatial domain as shown in Fig.\ref{fig: continuation}.
The computation of $f_{sdf}^{*}$ at $\bm{x}_{ob}$ with respect to $\mathcal{SV}_{\mathcal{B}(t)}$ yields a new implicit function. This involves identifying the argmin $t^*$ that corresponds to the minima of the implicit value, evaluated relative to the motion of the body $\mathcal{B}$ along the entire trajectory.
 
 \subsection{Space-time Continuation Efficient SDF Computation}
\label{sec: continuous-sdf}

We compute $t^*$ in equation (\ref{arg}) by a combination of gradient descent and Armijo line search \cite{armijo1966minimization} method.
The convergence speed of the algorithm depends on the choice of the initial value $t_{init}$. 
We perform trajectory sampling at uniform time intervals, calculate the shortest distance from $\bm{x}_{ob}$ to the sampled robot $\mathcal{B}$ along the trajectory, and obtain the corresponding moment as $t_{init}$, which is closed to $t^*$.
In addition,when calculating $f_{sdf}^{*}$ at several neighboring query points near $\bm{x}_{ob}$, we use $t^*$ of $\bm{x}_{ob}$ as $t_{init}$ . These strategies substantially improve the efficiency of the computation, making each query take only microseconds.
The complete computation can be stated as Algorithm \ref{alg1}.
           
\begin{algorithm}[!bh]
\caption{Efficient Computation for argmin and SDF}
\label{alg1}
\begin{algorithmic}[1]
\Require  Implicit function $\mathcal{SDF}^\mathcal{B}$, $R(t),\dot R(t),x(t),v(t)$ \\
\textbf{Initialize:} $t \leftarrow t_{init}, x \leftarrow \bm{x}_{ob}, \eta \leftarrow 0.02, c \leftarrow 0.5$ 
\Procedure {Propagate}{$t$} \label{process} \Comment{line 2} \label{process}
\State search direction: $d \leftarrow - {\dot f_{s d f}}\big|_{x}^{t^k}$ \Comment{using \ref{argdot}} 
\While{$f_{sdf}(x,t^k+\eta d)>f_{sdf}(x,t^k)+c \eta  d{\dot f_{s d f}}\big|_{x}^{t^k} $} \Comment{using \ref{SDFquery2}}
\State $\eta \leftarrow \eta/2$
\EndWhile\label{linesearch}
\State update iterate $t^{k+1} \leftarrow t^{k}+\eta d$
\State \Goto{process} unless the condition of convergence or termination is satisfied 
\State get $t^*$ and corresponding $f_{sdf}^{*}$
\EndProcedure
\end{algorithmic}
\end{algorithm}

\section{Optimization-Based Trajectory Generation}
\label{sec: Trajectory Generation}
We use whole-body planning of quadrotors as a case study here. 
Quadrotors have the property of differential flatness, which allows the attitude trajectory to be derived from the position trajectory, thus reducing the dimensionality of the trajectory optimization problem \cite{faessler2017differential}.

\subsection{Trajectory Representation}
\label{subsec: Trajectory Representation}
In this work, we adopt $\mathfrak{T}_{\textbf{MINCO}}$\cite{wang2022geometrically} to represent trajectories, which is a minimum control effort polynomial trajectory class defined as:
\setcounter{MaxMatrixCols}{20}
\begin{align}
\begin{split}
\mathfrak{T}_{\textbf{MINCO}}=&\{p(t):[0,T_\Sigma]\rightarrow\mathbb{R}^m|\textbf{c}=\mathcal{M}(\textbf{q},\textbf{T}),\\
\textbf{q}\in&\mathbb{R}^{(M-1)m},\textbf{T}\in\mathbb{R}^M_{>0} \},  \\
\textbf{c}=&(\textbf{c}^T_1,...,\textbf{c}^T_M)^T\in\mathbb{R}^{2Ms\times m}, \\
\textbf{q}=&(\textbf{q}_1,...,\textbf{q}_{M-1})\in\mathbb{R}^{(M-1)\times m}, \\
\textbf{T}=&(T_1,T_2,...,T_M)^T\in\mathbb{R}^{M}\}.\\
\end{split}
\end{align}
The trajectory $p(t)$ is an $m$-dimensional polynomial with $M$ pieces and degree $N=2s-1$, where $s$ is the order of the relevant integrator chain. $\textbf{c}$ is polynomial coefficients and $\textbf{q}$ is intermediate waypoints. The time allocated for each piece is given by $\textbf{T}$, and the total time is $T_\Sigma=\sum_{i=1}^MT_i$. The parameter mapping $\mathcal{M}(\textbf{q},\textbf{T})$ is constructed based on Theorem 2 in \cite{wang2022geometrically}.

An $m$-dimensional $M$-segment trajectory is described by the function as:
\begin{align}
p(t)=p_i(t-t_{i-1}) \quad \forall t \in [t_{i-1},t_i),
\end{align}
where the $i_{th}$ segment of the trajectory is represented by a polynomial of degree $N=5$:
\begin{align}
p_i(t)=\textbf{c}^T_i\beta(t) \quad \forall t \in [0,T_i).
\end{align}
$\textbf{c}_i \in \mathbb{R}^{(N+1)\times m}$ is the coefficient matrix, $\beta(t)=[1,t,...,t^N]^T$ is the natural basis, and $T_i=t_i-t_{i-1}$ is the time duration of the $i^{th}$ segment.
The trajectory representation $\mathfrak{T}_{\textbf{MINCO}}$ is uniquely determined by the pair $(\textbf{q},\textbf{T})$. The mapping $\textbf{c}=\mathcal{M}(\textbf{q},\textbf{T})$ converts the representation $(\textbf{q},\textbf{T})$ into $(\textbf{c},\textbf{T})$, allowing any second-order continuous cost function $J(\textbf{c},\textbf{T})$ to be expressed as $H(\textbf{q},\textbf{T})=J(\mathcal{M}(\textbf{q},\textbf{T}),\textbf{T})$. As a result, the partial derivatives $\partial H/\partial \textbf{q}$ and $\partial H/\partial \textbf{T}$ can be obtained from $\partial J/\partial \textbf{c}$ and $\partial J/\partial \textbf{T}$ with ease.

\subsection{Optimization Problem Formulation}
\label{subsec: Optimization Problem Formulation}
In this paper, we focus on trajectory generation for robots with quadrotor dynamics. To summarize,
trajectory generation can be constructed as the following unconstrained optimization problem:
\begin{align}
&\underset{\textbf{c},\textbf{T}}{{\min} } \  \lambda_s J_s+\lambda_m J_m+\lambda_d J_d+\rho J_t, 
\end{align}
where the terms $J_s, J_m, J_d, J_t$ are the safety, smoothness, dynamic feasibility, and total time penalties respectively. $\lambda_s, \lambda_m, \lambda_d$ and $\rho$ are their corresponding weights.

Typically, the safety penalty term $J_s$ for optimization is a safety evaluation integral over the entire trajectory, e.g. $J_s = \int_{t_{min}}^{t_{max}} J_s(\textbf{c}, \textbf{T}, t)\, dt$. Since the integral result has no analytical form, it is often approximated by discrete summation in practical applications. However, obtaining safety penalties by discrete sampling along the trajectory poses a risk of missing collisions between sampled instants, leading to the occurrence of the tunneling phenomenon \cite{ericson2004real}. Moreover, in a sparse environment, many sampling points are safe enough, evaluating these points is thus redundant, which reduces efficiency. In contrast,
our approach does not require sampling along the trajectory. Due to the properties of the signed distance to the swept volume, we only need to evaluate the corresponding $f_{sdf}^*$ at obstacle points. This approach theoretically avoids the tunneling phenomenon and has higher efficiency.

\subsubsection{Safty Penalty $J_s$ }
\label{subsubsec: Safty Penalty}
The purpose of the trajectory optimization is to ensure that the swept volume completely avoids obstacles, i.e. that the corresponding $f_{sdf}^*$ at all obstacle points are greater than a safety margin $s_{thr}$. Therefore, we construct a safety penalty using $t^*$ and $f_{sdf}^*$ derived from \ref{sec: continuous-sdf} to deform our trajectory. The penalty function is:
\begin{align}
&J_s=\sum_{i=1}^{N_{obs}}\mathcal{C}\left(\mathcal{G}_s(\bm{x}_{ob}^i)\right),\\
\label{gradient3}
&\mathcal{G}_s(\bm{x}_{ob})=
\begin{cases}
0, & \text{$ f_{sdf}^*(\bm{x}_{ob}) > s_{thr} $},\\
s_{thr}-f_{sdf}^*(\bm{x}_{ob}), & \text{$ f_{sdf}^*(\bm{x}_{ob})\le s_{thr} $},\\
\end{cases}\\
&  f_{s d f}^{*} \left( \bm{x}_{ob}\right)\! = \!\mathcal{SDF}^{\mathcal{B}}\left(R^{-1}(t^*)(\bm{x}_{ob}-p(t^*))\right), \\
&p(t^*)=c^T_l\beta(t^*\!-\!T_0\!-\!T_1\cdots\! -\! T_{l-1})\quad \text{located at $l_{th}$ piece},
\end{align}
 where $s_{thr}$ is the safety threshold. $\bm{x}_{ob}$ is the obstacle point near the trajectory selected by the Axis-aligned Bounding Box (AABB) algorithm and $N_{obs}$ is the number of  points selected.  $\mathcal{C}(\cdot)=\max\{\cdot,0\}^3$ is the cubic penalty. 

 The gradients are:
\begin{align}
&\frac{\partial J_s}{\partial {\textbf{c}}}=3\sum_{i=1}^{N_{obs}}\mathcal{Q}\left(\mathcal{G}_s(\bm{x}_{ob}^i)\right)\cdot {\left(\frac{\partial \mathcal{G}_s(\bm{x}_{ob}^i)}{\partial {\textbf{c}}}\right)}\bigg|_{t^*_i},\\
&\frac{\partial J_s}{\partial \textbf{T}}=3\sum_{i=1}^{N_{obs}}\mathcal{Q}\left(\mathcal{G}_s(\bm{x}_{ob}^i)\right)\cdot {\left(\frac{\partial \mathcal{G}_s(\bm{x}_{ob}^i)}{\partial \textbf{T}}\right)}\bigg|_{t^*_i}, \\
&\frac{\partial\mathcal{G}_s(\bm{x}_{ob})}{\partial \textbf{c},\textbf{T}} = 
\begin{cases}
0, & \text{$ f_{sdf}^*(\bm{x}_{ob}) > s_{thr} $},\\
-\frac{\partial f_{sdf}^*(\bm{x}_{ob})}{\partial \textbf{c}, \textbf{T}}, & \text{$ f_{sdf}^*(\bm{x}_{ob})\le s_{thr} $},\\
\end{cases}\\
\label{gradient4}
\end{align}
where $\mathcal{Q}(\cdot)=\max\{\cdot,0\}^2$ is the quadratic penalty.

For optimization purposes, we use a normalized quaternion, represented by $\textbf{q}=[w \, x \, y\, z]^{T}$ to denote the rotation. The corresponding rotation matrix $R$, is given as:
\begin{equation}
\label{rotate}
    R\!=\!\left[\begin{array}{ccc}
1\!-2\!\left(y^2\!+\!z^2\right) & 2(x y\!-\!w z) & 2(x z\!+\!w y) \\
2(x y\!+\!w z) & 1\!-\!2\!\left(x^2\!+\!z^2\right) & 2(y z\!-\!w x) \\
2(x z\!-\!w y) & 2(y z\!+\!w x) & 1\!-\!2\!\left(x^2\!+\!y^2\right)
\end{array}\right].
\end{equation}
Recall that $R^{-1}(t)\!=\!R(t)^{T}$. Given this property, the partial derivatives of $R^{-1}(t)$ and $R(t)$ with respect to $\textbf{q}.*$ can be easily obtained, where $*$ denotes the elements $w\,x\,y\,z$ in quaternions.
Differentiating the equation (\ref{SDFquery2}) with respect to $p(t)$ and $\textbf{q}(t)$ gives the gradients of the signed distance evaluated at $\bm{x}_{ob}$ with respect to rotations and translations:

\begin{align}
\label{gradient1}
& {\frac{\partial f_{sdf}(\bm{x}_{ob})}{\partial p}=-(\nabla \mathcal{SDF}^{\mathcal{B}}}\big|_{\bm{x}_{rel}})^T \cdot R^{-1}(t), \\
& {\frac{\partial f_{sdf}(\bm{x}_{ob})}{\partial \textbf{q}.*}=(\nabla \mathcal{SDF}^{\mathcal{B}}}\big|_{\bm{x}_{rel}})^T \cdot \frac{\partial R^{-1}(t)}{\textbf{q}.*}\cdot (\bm{x}_{ob}-p(t)), 
\end{align}
where $\nabla{\mathcal{SDF}^{\mathcal{B}}}\big|_{\bm{x}_{rel}}$ is the gradient of the SDF of the robot body $\mathcal{B}$ at the point $ \bm{x}_{rel} $.

\begin{figure}[!t]                              
\centering
\vspace{0.0cm}
{\includegraphics[width=0.97\columnwidth]{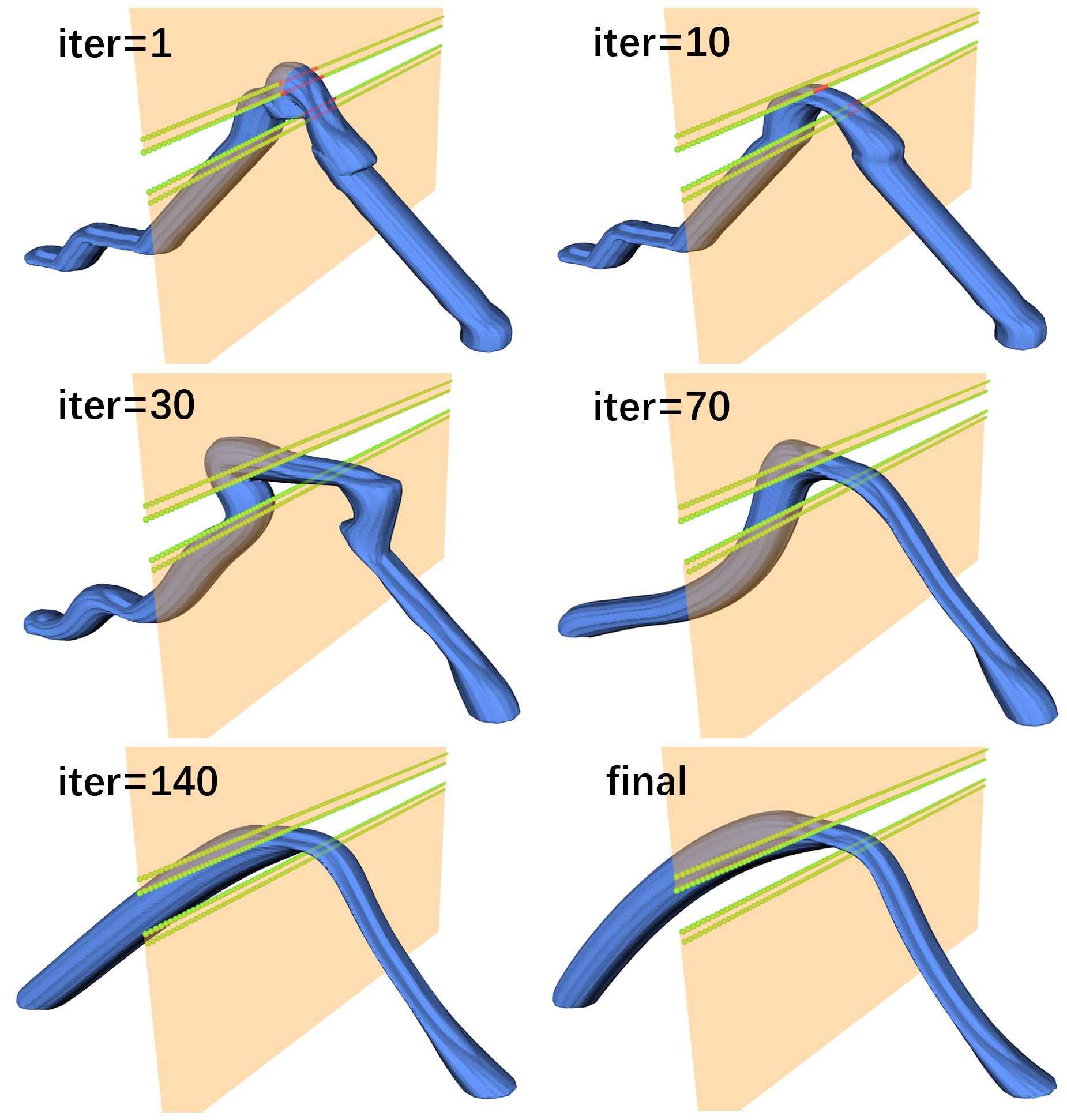}}
\captionsetup{font={small}}
\caption{ \label{fig:iter}This figure shows a trajectory snapshot along with its swept volume during the optimization process. A UFO robot manages to fly through narrow gaps, requiring whole-body planning. Red dots represent obstacles that do not satisfy safety constraints.}
\vspace{-1.5cm}
\end{figure}

\begin{figure*}[!t]  
\vspace{-0.25cm}  
\centering
{\includegraphics[width=2.0\columnwidth]{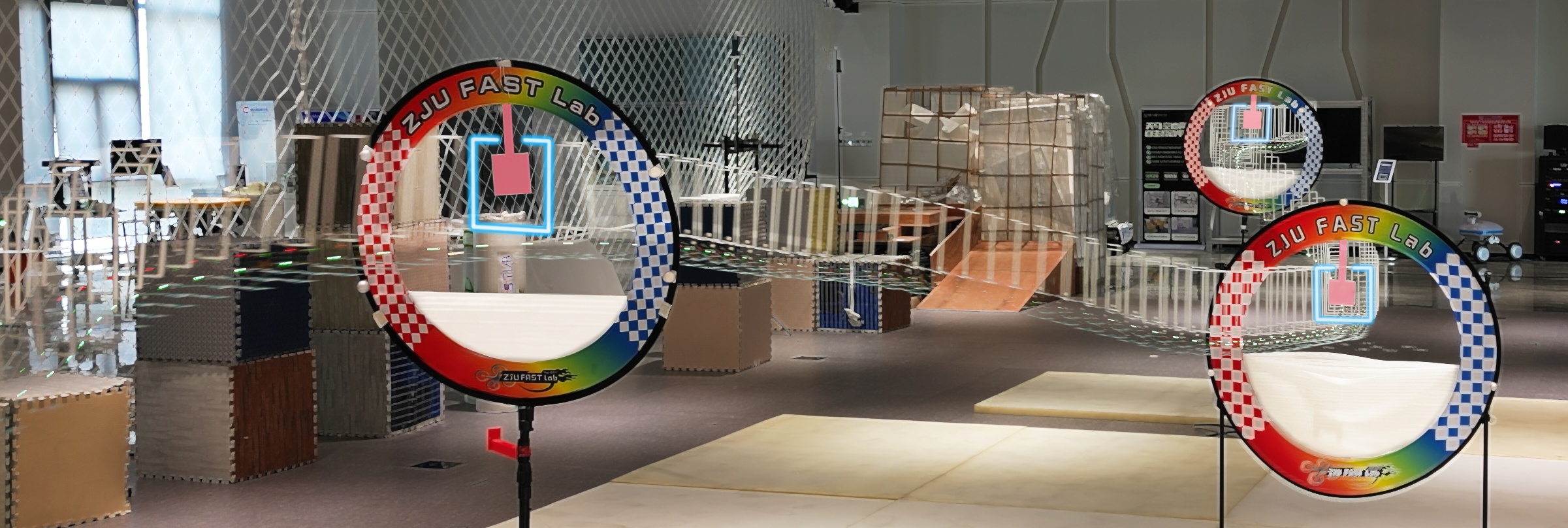}}
\captionsetup{font={small}}
\caption{ \label{fig:real1_1}The real-world indoor experiment: The quadrotor must traverse three consecutive circles  while precisely avoiding nearby obstacles. 
We highlight the most critical frames when the robot is closest to the obstacles.  }
\vspace{-0.10cm}  
\label{fig:real_exp1}
\end{figure*}
\begin{figure*}[!th]  
\vspace{-0.00cm}  
\centering
{\includegraphics[width=2.0\columnwidth]{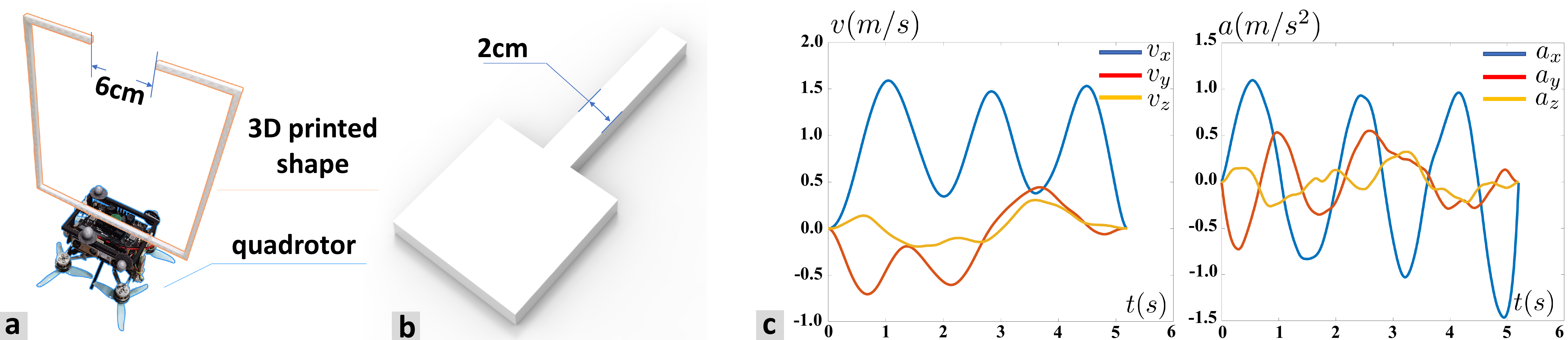}}
\captionsetup{font={small}}
\caption{ \label{fig:real1_2}The UAV platform and the obstacle in the experiment are shown in Fig.a and Fig.b.The velocity and acceleration of the drone in the real experiment are shown in Fig.c. }
\label{fig:real_exp2}
\vspace{-0.3cm}  
\end{figure*}

By choosing $s=3$ as the integrator chain and using the differential flatness property of quadrotors, it is possible to propagate the gradients with respect to rotation $\textbf{q}.*$ to position, velocity, acceleration, and jerk.  Furthermore, the gradients of the signed distance evaluated at $\bm{x}_{ob}$ with respect to $\textbf{c}_k, \textbf{T}_k$ can be derived as follows, where $\zeta$ denotes position, velocity, acceleration, and jerk, respectively.
\begin{align}
\label{gradient2}
& \frac{\partial f_{sdf}^*(\bm{x}_{ob})}{\partial \textbf{c}_k}=\sum_{\zeta=p,\,v,\,a\,j}{}\frac{\partial f_{sdf}^*(\bm{x}_{ob})}{\partial \zeta} \cdot \frac{\partial \zeta(t)}{\partial \textbf{c}_k}, \\
& \frac{\partial f_{sdf}^*(\bm{x}_{ob})}{\partial \textbf{T}_k}=\sum_{\zeta =p,\,v,\,a\,j}{}\frac{\partial f_{sdf}^*(\bm{x}_{ob})}{\partial \zeta} \cdot \frac{\partial \zeta(t)}{\partial \textbf{T}_k}.
\end{align}

Note that the trajectory deformation causes $t^*$ to be different for the same $\bm{{x}}_{ob}$. 
Therefore, $ t^* $ is also a function of the optimization variables $ \textbf{c}, \textbf{T} $.
However, since $t^*$ is obtained from an optimization problem, it is difficult to explicitly derive the derivatives of $t^*$ with respect to $\textbf{c}, \textbf{T}$. 
Fortunately, sufficient conditions for optimality for deriving $t^*$ allow us to obtain these derivatives implicitly.
In essence, in the equation (\ref{argdot}), $\dot{f}_{sdf}\equiv 0 $ when $t = t^*$. By utilizing this identity equation, the required gradients can be derived. \cite{gould2016differentiating} has a description of this method. Implementation details can be found in Appendix \ref{appendix}.

\subsubsection{Smoothness Penalty $J_m$ }
\label{subsubsec: Smoothness Penalty}
To ensure the smoothness of the trajectory, we minimize the integral of the third-order derivative of the trajectory:
\begin{align}
&J_m=\int_{0}^{T_{\Sigma}} j^2(t) dt,
\end{align}
where $T_{\Sigma} = \sum^M_{i=1}T_i$ is the total time, $j(t)$ denotes jerk. The gradients are:
\begin{align}
    \frac{\partial J_m}{\partial \textbf{c},\textbf{T}} = 2\int_{0}^{T_{\Sigma}} j(t) \frac{\partial j(t)}{\partial \textbf{c},\textbf{T}} dt.
\end{align}
\subsubsection{Dynamical Feasibility Penalty $J_d$ }
\label{subsubsec: Smoothness and Dynamical Feasibility Penalty}
To satisfy the dynamic constraints of the robot, we limit the maximum velocity and thrust:
\begin{align}
&J_d=\int_{0}^{T_{\Sigma}}\mathcal{C}\left(\mathcal{G}_d(\xi(t))  \right) dt,\\
&\mathcal{G}_d(\xi(t))=
\begin{cases}
0, & \text{$ \xi \le \xi_{max} $},\\
\xi - \xi_{max}, & \text{$ \xi > \xi_{max} $},\\
\end{cases}
\end{align}
where $\xi$ denotes velocity and thrust respectively.
The gradients are:
\begin{align}
    \frac{\partial J_d}{\partial \textbf{c},\textbf{T}} = 3\int_{0}^{T_{\Sigma}} \mathcal{Q}\left(\mathcal{G}_d(\xi(t))  \right) \frac{\partial \xi(t)}{\partial \textbf{c},\textbf{T}} dt.
\end{align}

\subsubsection{Total Time Penalty $J_t$ }
\label{subsubsec: Total Time Penalty}
We minimize the total time $J_t=\sum^M_{i=1}T_i$ to improve the aggressiveness of the trajectory, the gradients are $\partial J_t / \partial\textbf{c}=\textbf{0}$ and $\partial J_t / \partial\textbf{T}=\textbf{1}$.


To solve this optimization problem, we use a numerical algorithm. Fig.\ref{fig:iter} shows the deformation of the trajectory for different iterations and the corresponding swept volume of the UFO robot.

\section{Results}
\subsection{Implementation details}
\label{subsec: Implementation details}
We validate the approach on a quadrotor platform. Some additional structures are mounted on the quadrotor to simulate a vehicle with complex shapes. All computations are performed on an onboard computer: Nvidia Xavier NX.

We choose the L-BFGS\footnote{\url{https://github.com/ZJU-FAST-Lab/LBFGS-Lite/}}  algorithm\cite{liu1989limited} as a highly efficient quasi-Newton approach for solving numerical optimization problems and use the Lewis-Overton line search \cite{lewis2013nonsmooth} to address instances of non-smoothness in the scale that may arise during the optimization process.
\subsection{The Real-World Experiment }

\label{subsec: Real-World Experiments}
We conduct a real-world experiment in an indoor environment. 
Fig.\ref{fig:real_exp2}a shows the quadrotor robot in our experiment. We deliberately install a structure on the robot to give it a complex shape. 
We hang the obstacle shown in Fig.\ref{fig:real_exp2}b inside a ring so that the robot has to cross the ring from the correct position to ensure that it does not collide. The maximum safe distance is only 2cm, thus testing the accuracy of our algorithm. In this experiment, the maximum speed and acceleration of the quadrotor are limited to $2m/s$ and $3m/s^2$ Fig.\ref{fig:real_exp1} and Fig.\ref{fig:real_exp2}c shows the result.

\subsection{Simulation Experiments }
\label{subsec: Simulation Experiment}
To further validate the capability of the proposed algorithm, we construct more complex environments for simulations with a variety of robot shapes. Similarly, a dynamic model of quadrotors is used for optimization.

\begin{figure}[htbp]  
\vspace{0.2cm}  
\centering
{\includegraphics[width=1.0\columnwidth]{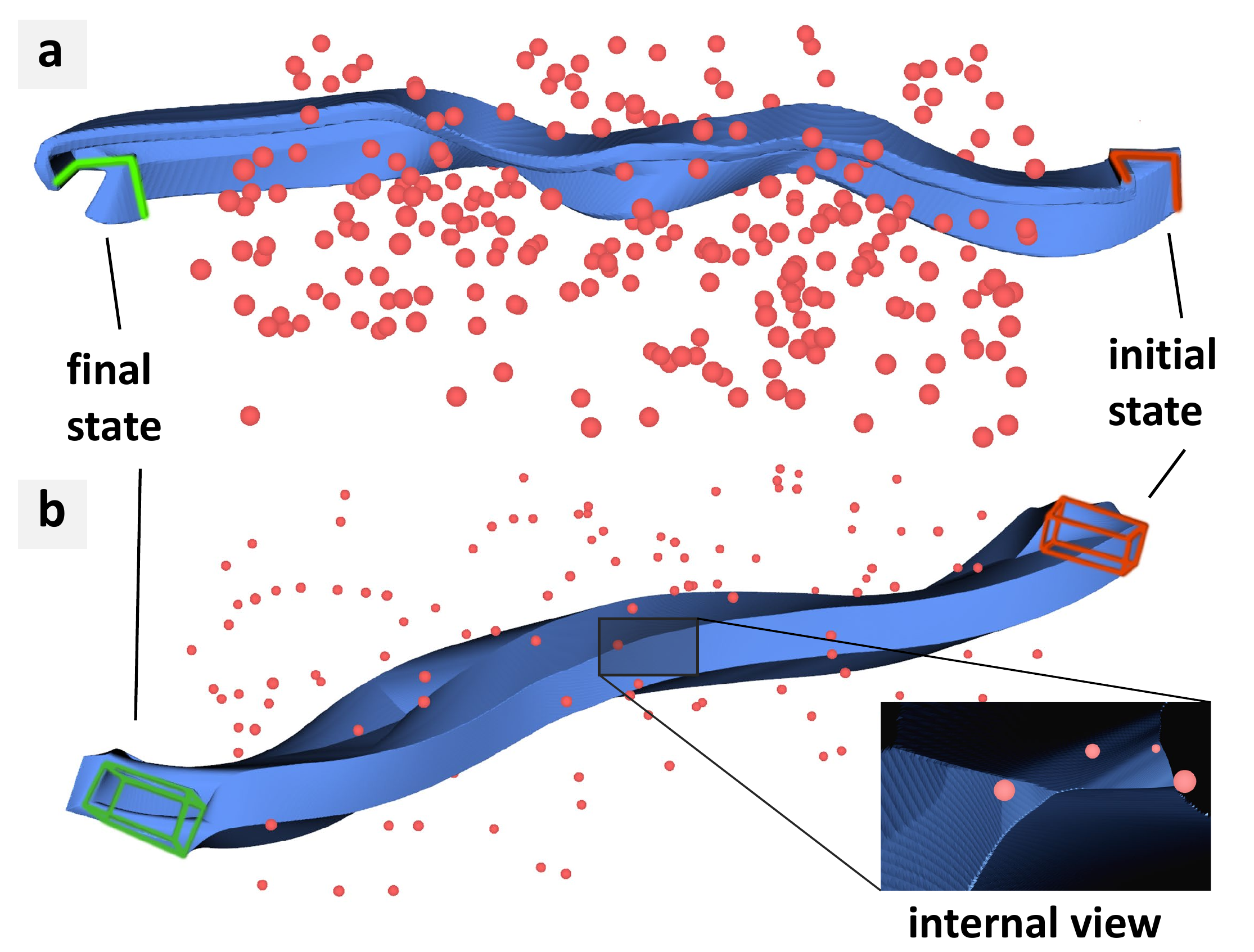}}
\captionsetup{font={small}}
\caption{ \label{fig:sim1_result} The results of the first simulation experiment. Pink spheres represent obstacles. Fig.a and Fig.b denote the swept volume of the optimized trajectory of $\mathcal{B}^{\mathcal{X}}$ and $\mathcal{B}^{\mathcal{Y}}$ separately. Since $\mathcal{B}^{\mathcal{Y}}$ is hollow, there are also some cavities within its swept volume. The close-up of Fig.b shows the obstacles inside these cavities, demonstrating that our algorithm can make full use of the feasible  space.}
\vspace{0.0cm}  
\end{figure}

\begin{figure}[htbp]  
\vspace{-0.2cm}  
\centering
{\includegraphics[width=0.97\columnwidth]{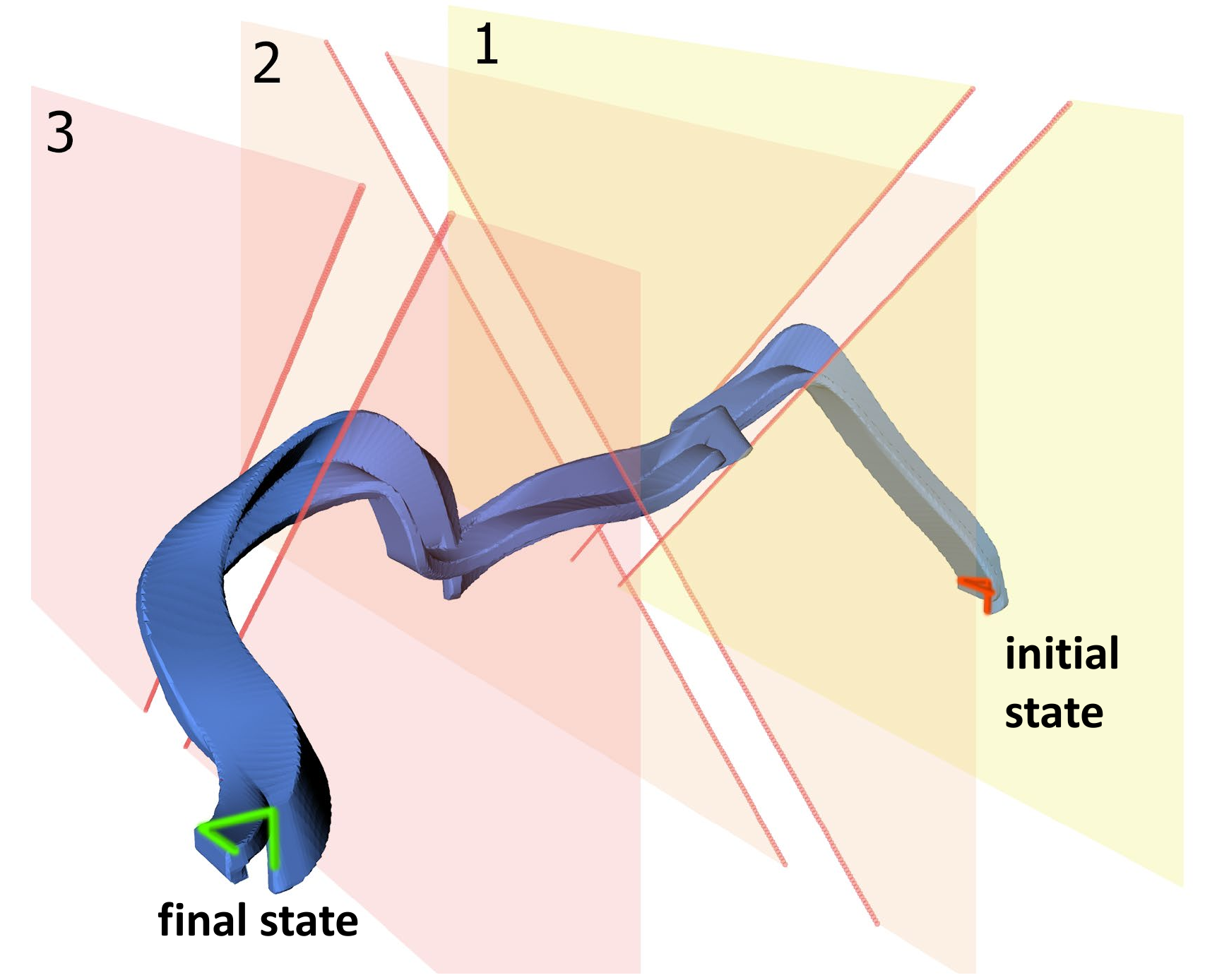}}
\captionsetup{font={small}}
\caption{ \label{fig:sim2_result} The result of the second simulation experiment. Since the gaps are narrow, the robot must perform a large attitude maneuver to get through them.}
\vspace{-0.0cm}  
\end{figure}

Two different shapes of robots denoted as $\mathcal{B}^{\mathcal{X}}$ and $\mathcal{B}^{\mathcal{Y}}$ are shown in Fig.\ref{fig:sim1}.
 In the first experiment, the environment consists of randomly generated dense obstacles, and the robots traverse this area separately. 
It is worth noting that our algorithm also works for robots with a hollow shape like $\mathcal{B}^{\mathcal{Y}}$. 
In the second experiment, the environment consists of three sloping narrow gaps, and $\mathcal{B}^{\mathcal{X}}$ traverses the three gaps in turn while avoiding collisions.
Fig.\ref{fig:sim1_result} and Fig.\ref{fig:sim2_result} show the swept volume corresponding to the final trajectory.
Due to the limitations of the visualization, we recommend  watching our video\footnote{\url{https://drive.google.com/file/d/1-QQZILtCd5WudsjGIY2Y1KUItSsiGuco/view}} for a more detailed view of the experimental result.

\section{Conclusion and Future Work}
\label{sec:Conclusion}
We present a novel approach to implicitly, lazily, and efficiently compute the signed distance to the swept volume constructed by a robot and its trajectory using the continuity in space-time. Furthermore, we also integrate the continuous implicit SDF into the whole-body optimization problem using quadrotors as a case study.

In principle, our methodology does not impose any restrictions on the class, shapes, or trajectories of robots. Taking advantage of the implicit representation and the analytic form of gradients, this method can also be implemented for any trajectory except polynomials, as long as it is differentiable with respect to time. It is also worth mentioning that we will consider time-variant deformable robots of any shape described by the implicit function. We will also validate this pipeline for different robots with different dynamics for the completeness of planning.

\section{Appendix}
\label{appendix}

According to the differntial flatness, there is $\dot{R}=R \hat{\omega}$ for quadrotors
 , we can simplify the equation (\ref{argdot}) as follows:
\begin{equation}
 \label{argdot2} 
  {\dot f_{s d f}}\big|_{\bm{x}_{ob}} =({\nabla{\mathcal{SDF}^{\mathcal{B}}}\big|_{\bm{x}_{rel}}})^{T}(\hat{\omega} R^{T}(p-\bm{x}_{ob})-R^{T} v).
 \end{equation}
For simplicity, we use some symbols to represent some formulas above as follows: 

\begin{figure}[htbp]  
\vspace{1.2cm}  
\centering
{\includegraphics[width=1.0\columnwidth]{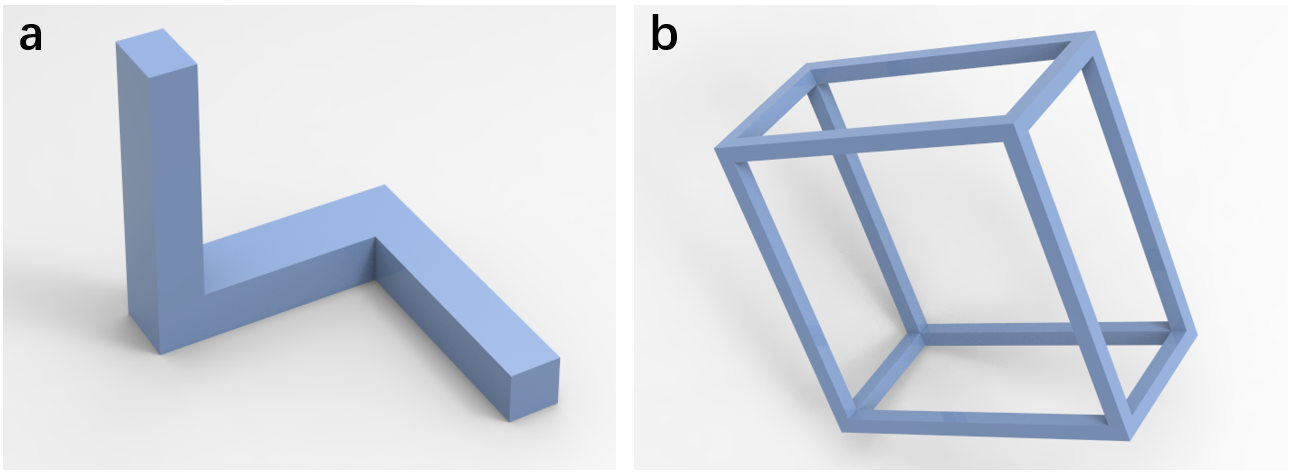}}
\captionsetup{font={small}}
\caption{ \label{fig:sim1} Two robots with complex shapes in simulation experiments. For convenience, we call the robot in Fig.a  $\mathcal{B}^{\mathcal{X}}$ and the robot in Fig.b $\mathcal{B}^{\mathcal{Y}}$. }
\end{figure}

 \begin{align}
 & \mathcal{X}(R,p) \triangleq \left({\nabla{\mathcal{SDF}^{\mathcal{B}}}\big|_{\bm{x}_{rel}}} \right)^T, \\ 
&  \mathcal{Y}(R,\hat{\omega},p,v) \!\triangleq\! \hat{\omega} R^{T}(p\!-\!\bm{x}_{ob})\!-\!R^{T} v,\\
& \mathcal{F}(t^*,\zeta) \triangleq {\dot f_{s d f}\big|_{\bm{x}_{ob}}} = \mathcal{X} \cdot \mathcal{Y} \equiv 0,
 \end{align}
where $\zeta$ denotes $p,v,\omega,R$.
Deriving the above equations we can obtain:

\begin{equation}
\begin{split}
        \frac{\partial \mathcal{X}}{\partial {t^*}}\!&=\! \!\left(\!{\nabla}^{2}\mathcal{SDF}^{\mathcal{B}}\big|_{\bm{x}_{rel}}\!\left(\!\frac{\partial R^{T}}{\partial {t}}(\bm{x}_{ob}-p)-R^Tv \!\right)\!\right)^T, \\
        \frac{\partial \mathcal{Y}}{\partial {t^*}}&= \frac{\partial \hat{\omega}}{\partial t}(R^Tp)+\hat{\omega}(\frac{\partial R^{T}}{\partial {t}}p+R^Tv) - R^{T}a \\
        & -\frac{\partial R^{T}}{\partial {t}}v - \hat{\omega}\frac{\partial R^{T}}{\partial {t}}\bm{x}_{ob}-\frac{\partial \hat{\omega}}{\partial t}R^T\bm{x}_{ob}.
\end{split}
\end{equation}
Recall that:
\begin{equation}
    \mathcal{F}(t^*(\zeta),\zeta) \equiv 0,
\end{equation}
\begin{equation}
\frac{\mathrm{d} \mathcal{F}}{\mathrm{d} \zeta} =\frac{\partial \mathcal{F}}{\partial t^*}\frac{\partial t^*}{\partial \zeta}+\frac{\partial \mathcal{F}}{\partial \zeta} \equiv 0.
\end{equation}
\begin{align}
    &\frac{\partial t^*}{\partial \zeta}=-\frac{\partial \mathcal{F}}{\partial \zeta}/\frac{\partial \mathcal{F}}{\partial t^*}, \\
    &\frac{\partial \mathcal{F}}{\partial t^*}=\mathcal{X} \frac{\partial \mathcal{Y}}{\partial {t^*}}+ \frac{\partial \mathcal{X}}{\partial {t^*}}\mathcal{Y}.
\end{align}
Finally, we can get the corresponding gradients associated with $t^*$ as follows:

$$ \mathcal{K} \triangleq (\mathcal{X} \frac{\partial \mathcal{Y}}{\partial {t^*}}+ \frac{\partial \mathcal{X}}{\partial {t^*}}\mathcal{Y}), $$
\begin{align}
\frac{\partial t^*}{\partial v}&=(\mathcal{X}R^T)/\mathcal{K},\\
\frac{\partial t^*}{\partial \omega}&=-\left(\mathcal{X}\frac{\partial \hat{\omega}}{\partial \omega}R^T(p-\bm{x}_{ob})\right)/\mathcal{K}, \\
\frac{\partial t^*}{\partial p}&=\left({\nabla}^{2}\mathcal{SDF}^{\mathcal{B}}\big|_{\bm{x}_{rel}}R^T\mathcal{Y}+\mathcal{X}\hat{\omega} R^T\right)/\mathcal{K},\\
\frac{\partial t^*}{\partial \textbf{q}.*}&= \left( {\nabla}^{2}\mathcal{SDF}^{\mathcal{B}}\big|_{\bm{x}_{rel}}\frac{\partial R^{T}(t)}
 {\textbf{q}.*} (\bm{x}_{ob}-p) \mathcal{Y} \right) / \mathcal{K} \nonumber \\
    &+\left( \mathcal{X}(\hat{\omega}\frac{\partial R^{T}(t)}{\textbf{q}.*}(p-\bm{x}_{ob})-v)) \right) / \mathcal{K}.
\end{align}

Similar to equation (\ref{gradient2}), the final gradient $\partial {t^*} / \partial {\textbf{c},\textbf{T}} $ can be obtained by the differential flatness property.

\newlength{\bibitemsep}\setlength{\bibitemsep}{0.00\baselineskip}
\newlength{\bibparskip}\setlength{\bibparskip}{0pt}
\let\oldthebibliography\thebibliography
\renewcommand\thebibliography[1]{
    \oldthebibliography{#1}
    \setlength{\parskip}{\bibitemsep}
    \setlength{\itemsep}{\bibparskip}
}
\bibliography{main}

\end{document}